\documentclass[a4paper,fleqn]{cas-dc}

\usepackage[authoryear,longnamesfirst]{natbib}
\usepackage{multirow}

\def\tsc#1{\csdef{#1}{\textsc{\lowercase{#1}}\xspace}}
\tsc{WGM}
\tsc{QE}
\tsc{EP}
\tsc{PMS}
\tsc{BEC}
\tsc{DE}

\begin{document}
\let\WriteBookmarks\relax
\def\floatpagepagefraction{1}
\def\textpagefraction{.001}

\shorttitle{OWL: Benchmarking Weakly Supervised Learning for Aerial Wildlife Surveys}

\shortauthors{Chacon et~al.}

\title [mode = title]{Overhead Wildlife Locator (OWL): Benchmarking Weakly Supervised Learning for Aerial Wildlife Surveys}



\author[1,2]{Isai Daniel Chac\'on}[]
\cormark[1]
\ead{v-ichaconsil@microsoft.com}

\author[1]{Zhongqi Miao}[]

\author[1]{Bruno Demuro}[]

\author[1]{Caleb Robinson}[]

\author[1]{Rahul Dodhia}[]

\author[3]{Lasha Otarashvili}[]

\author[3]{Jason Holmberg}[]

\author[4]{Kirk Larsen}[]

\author[5]{Howard Frederick}[]

\author[6]{Nathan J. Pamperin}[]

\author[2]{Pablo Arbel\'aez}[]

\author[1]{Juan M. Lavista Ferres}[]


\affiliation[1]{organization={Microsoft AI for Good Lab},
    city={Redmond},
    state={WA},
    country={USA}}

\affiliation[2]{organization={Center for Research and Formation in Artificial Intelligence (Cinfonia), Universidad de los Andes},
    city={Bogot\'a},
    country={Colombia}}

\affiliation[3]{organization={Conservation X Labs},
    city={Washington, D.C.},
    country={USA}}

\affiliation[4]{organization={Kirk Larsen Consulting},
    city={Langley},
    state={WA},
    country={USA}}

\affiliation[5]{organization={Tanzania Wildlife Research Institute},
    city={Arusha},
    country={Tanzania}}

\affiliation[6]{organization={Alaska Department of Fish and Game},
    city={Juneau},
    state={AK},
    country={USA}}

\begin{abstract}
Automated aerial wildlife surveys increasingly rely on deep learning, yet
standard object detectors require bounding-box annotations, reported to be up
to seven times slower and three times more expensive to produce than
point-level labels. To address this bottleneck, we introduce the Overhead
Wildlife Locator (OWL), a weakly supervised density-estimation framework with
three variants: OWL-C, a fully convolutional model for high-throughput
screening; OWL-T, a Swin-augmented hybrid for heterogeneous, cluttered scenes;
and OWL-D, built on a frozen DINOv3 ViT-H+/16 encoder with a DPT-style fusion
decoder. We benchmark all three against POLO, YOLOv11n, and YOLOv11l
across five public aerial datasets --- from sparse fixed-wing savanna surveys to
dense UAV paddock imagery --- and against the published HerdNet baseline on its
native Delplanque split.
OWL-D sets a new state of the art on Delplanque ($0.934$ AP vs.\ HerdNet's $0.840$) and records the highest AP on four of the five datasets. Performance is regime-dependent: on the extreme-density SheepCounter UAV dataset the hybrid OWL-T leads ($0.978$ AP) and the convolutional variants attain the lowest counting error, whereas the foundation-based OWL-D degrades --- indicating which variant suits which survey type.
We further validate operational readiness on the Alaska Department of Fish and
Game's 2022 Central Arctic Caribou census: under cross-herd and cross-temporal
transfer, OWL-C fine-tuned on the 2017 Porcupine Caribou Herd split attains
$F_1 = 0.965$ on a held-out patch test set, with a signed count error
of $+3.1\%$ aggregated across the released test patches. We release the OWL code, model weights, and the annotated Porcupine
Caribou Herd 2017 (PCH) and Central Arctic Herd 2022 (CAH) patches --- the first
open patch-level datasets for large-scale caribou aerial surveys --- at
\url{https://github.com/microsoft/MegaDetector-Overhead}.
\end{abstract}


\maketitle

\hypersetup{
  pdftitle={Overhead Wildlife Locator (OWL): Benchmarking Weakly Supervised Learning for Aerial Wildlife Surveys},
  pdfauthor={Isai Daniel Chacon; Zhongqi Miao; Bruno Demuro; Caleb Robinson; Rahul Dodhia; Lasha Otarashvili; Jason Holmberg; Kirk Larsen; Howard Frederick; Nathan J. Pamperin; Pablo Arbelaez; Juan M. Lavista Ferres},
  pdfsubject={Weakly supervised point-based density estimation for aerial wildlife localization and counting},
  pdfkeywords={Aerial Wildlife Monitoring; Density Estimation; Vision Transformer; Foundation Models; Point Annotation; Benchmark}
}

\section{Introduction}
Wildlife population monitoring has traditionally relied on manual counts performed by human observers from light aircraft \citep{norton1978counting, bayliss1989distribution, couturier1996calving}. These counts are prone to underestimation bias and high inter-observer variability at scale \citep{michaud2014estimating, stapleton2016aerial, fleming2008some, schlossberg2016testing}. High-resolution aerial imagery combined with deep learning has largely automated the analysis step \citep{lamprey2020cameras, tuia2022perspectives, kellenberger2018detecting, xu2024review}, but scalability is now constrained by an annotation bottleneck: standard object detectors such as the YOLO family \citep{redmon2016you} and Faster R-CNN \citep{ren2015faster} require bounding-box labels reported to be up to \textbf{seven times slower} to collect \citep{ge2023point} and roughly \textbf{three times more expensive} \citep{mullen2019comparing} than point-level annotations.

Point-level labelling --- marking only the geometric centre of each individual --- avoids the need to delineate spatial extent, a step that is especially ambiguous when targets are small, clustered, or captured in oblique imagery \citep{delplanque2023herdnet}. \citet{xu2025bounding} reported that point supervision attains statistically equivalent detection accuracy to bounding-box supervision in aerial surveys, and \citet{may2025minimize} showed that counting accuracy is largely preserved, with at most a $2.5\times$ MAE increase for a $7\times$ reduction in annotation effort. Together these results suggest that the additional cost of bounding-box delineation yields diminishing returns for surveys scoped to population counting and individual localization.

We introduce the \textbf{Overhead Wildlife Locator (OWL)} to address this gap. OWL extends the density-estimation framework of HerdNet \citep{delplanque2023herdnet} with a modular feature-extraction design and three variants that span the encoder-architecture spectrum: OWL-C, a fully convolutional model; OWL-T, a Swin-augmented hybrid \citep{liu2021swin}; and OWL-D, which replaces the CNN backbone with a frozen DINOv3 \citep{simeoni2025dinov3} ViT-H+/16 foundation encoder paired with a DPT-style \citep{ranftl2021vision} decoder. Section~\ref{sec:method} details each variant.

We benchmark OWL against POLO and the YOLOv11 family across five public aerial datasets spanning species, platforms, and density regimes --- from sparse fixed-wing savanna imagery to dense UAV paddock scenes --- and against HerdNet on its native Delplanque split. Operational readiness is further validated through a deployment with the Alaska Department of Fish and Game on the 2022 Central Arctic Caribou census (Section~\ref{subsec:caribou_case_study}). Our contributions are threefold:
\begin{enumerate}
    \item \textbf{Three architectural design points.} We propose OWL and study three feature-extraction strategies spanning the encoder spectrum: an ImageNet-pretrained supervised CNN encoder (OWL-C), Swin refinement of that encoder (OWL-T), and a frozen self-supervised foundation encoder paired with a DPT-style decoder (OWL-D). OWL-C and OWL-T share a backbone and decoder and differ only in the injected Swin blocks, so their contrast is a controlled comparison; OWL-D changes both the encoder and the decoder (DPT fusion vs.\ DLAUp) and is therefore a distinct design point rather than a single-variable encoder swap.

    \item \textbf{Unified benchmark.} We consolidate five aerial wildlife datasets into a standardized evaluation with bootstrap 95\% confidence intervals on every reported metric, comparing the three OWL variants against POLO, YOLOv11n, and YOLOv11l on all five datasets and against the published HerdNet baseline on its native Delplanque split (Section~\ref{subsec:implementation}).

    \item \textbf{Real-world validation and public dataset.} We deploy OWL with the Alaska Department of Fish and Game and publicly release the annotated Porcupine Caribou Herd (PCH) training patches and Central Arctic Herd (CAH) testing patches --- the first open patch-level datasets for large-scale caribou aerial surveys.
\end{enumerate}

\section{Related Work}
\label{sec:related}

\subsection{Deep Learning for Aerial Wildlife Monitoring}
Early work showed that convolutional neural networks (CNNs) could reliably detect large mammals in UAV and fixed-wing imagery across diverse habitats \citep{kellenberger2018detecting, eikelboom2019improving, linchant2015unmanned}. Subsequent reviews have catalogued the expansion of this field from species-specific detectors to multi-species models that operate across platforms \citep{tuia2022perspectives, xu2024review}. Detection architectures have converged on two paradigms: two-stage detectors such as Faster R-CNN \citep{ren2015faster}, which prioritize localization precision, and single-stage detectors such as YOLO \citep{redmon2016you}, preferred for high-throughput screening. Both paradigms require bounding-box supervision, which is ambiguous when animals are small, densely clustered, or partially obscured, and does not scale to the high-resolution imagery characteristic of modern census operations \citep{lamprey2020cameras}.

\subsection{Weakly Supervised Point-based Detection}
To overcome the annotation burden, point-level supervision has emerged as a practical alternative. Early density-estimation methods \citep{lempitsky2010learning} map pixels to object counts via Gaussian kernels but produce spatially blurred representations that merge adjacent animals in dense scenes, preventing precise localization.

HerdNet \citep{delplanque2023herdnet} addresses this by replacing Gaussian kernels with the Focal Inverse Distance Transform (FIDT), which generates sharp activation peaks at each animal centre and enables coordinate recovery via local-maxima detection. The approach is effective for species inhabiting homogeneous landscapes where animal--background contrast is low \citep{serati2026weakly}. Its primary limitation is the multi-task design: HerdNet jointly optimizes for density estimation and species classification, concurrent objectives that may introduce gradient conflicts and reduce localization precision relative to a task-specialized model. Its DLA-34 encoder is ImageNet-pretrained and then fine-tuned in-domain, rather than pretrained on a large self-supervised external corpus (Section~\ref{subsec:foundation-related}), bounding the representational quality available to the localization head.

Concurrently, anchor-free detectors have been adapted for point supervision. POLO \citep{may2024polo}, built on YOLOv8, treats point annotations as centres of pseudo-boxes with a fixed user-defined radius. A subsequent study \citep{may2025minimize} benchmarked this approach and found that pseudo-box detection outperforms localization on nadir images but that localization is preferable at high density or low pixel occupancy. The pseudo-box paradigm's central limitation is the fixed-radius assumption: when animal scale varies with altitude or clustering, overlapping pseudo-boxes trigger non-maximum suppression, discarding valid detections and undercounting dense herds.

\subsection{Vision Transformers and Global Context}
Although deep CNNs have large theoretical receptive fields, their effective receptive fields are substantially smaller and concentrated near the centre of the kernel stack \citep{luo2016understanding} --- a limitation that impairs the separation of small animals from visually similar background elements in complex terrain. Self-attention computes pairwise feature interactions across the entire input and addresses this gap. The Vision Transformer (ViT) \citep{dosovitskiy2020image} and its hierarchical variant, the Swin Transformer \citep{liu2021swin}, perform strongly on dense prediction in high-resolution aerial imagery \citep{aleissaee2022transformers}. Full self-attention on the high-resolution feature maps typical of aerial wildlife models is, however, computationally prohibitive. SwinIR \citep{liang2021swinir} showed that Residual Swin Transformer Blocks can be embedded within a CNN pipeline without this cost --- an architectural strategy we adapt to selectively inject global context into the feature pyramid for background suppression.

\subsection{Self-Supervised Foundation Models for Dense Prediction}
\label{subsec:foundation-related}
Vision Transformers trained without labels at large scale produce general-purpose dense feature representations that transfer broadly to downstream tasks. The DINO family of self-distillation methods \citep{caron2021emerging} established that self-supervised ViTs spontaneously develop object-aware attention maps; DINOv2 \citep{oquab2024dinov2} scaled this paradigm to curated billion-image corpora, and DINOv3 \citep{simeoni2025dinov3} introduced a Gram-matrix regularizer that preserves the spatial coherence of these features at high resolutions and across extended pretraining schedules. Translating patch-level token representations into the multi-scale feature pyramids expected by dense decoders requires architectural adapters: the Dense Prediction Transformer (DPT) of \citet{ranftl2021vision} reassembles ViT tokens into spatially rescaled feature maps and merges them through a bottom-up convolutional fusion stack, and ViT-Adapter \citep{chen2022vitadapter} injects spatial priors back into a frozen ViT. We adopt this paradigm in OWL-D. To our knowledge, OWL-D is the first system to couple a frozen DINOv3 encoder with a DPT-style decoder for FIDT-supervised wildlife localization in aerial imagery.

\section{Methodology}
\label{sec:method}

We formulate aerial wildlife localization as a weakly supervised density-estimation problem. Given an input image $I \in \mathbb{R}^{H \times W \times 3}$, we predict a density map $\hat{Y} \in [0,1]^{H/d \times W/d}$ with downsampling factor $d$; individual animal locations are recovered from $\hat{Y}$ via a Local Maxima Detection Strategy (LMDS, Section~\ref{subsec:inference}), and the count is the number of extracted peaks. OWL has three variants --- OWL-C, OWL-T, and OWL-D --- detailed in Section~\ref{subsec:architecture}.

\subsection{Ground Truth Generation: Focal Inverse Distance Transform}
\label{subsec:fidt}

Fixed-variance Gaussian kernels \citep{lempitsky2010learning} produce blurred density representations that merge adjacent objects. We instead adopt the Focal Inverse Distance Transform (FIDT) of \citet{liang2022focal}, which gives a sharper supervision signal and keeps individual animals within a herd distinguishable.

For a set of annotated point coordinates $P = \{p_1, p_2, \dots, p_N\}$ we compute the Euclidean distance transform $D(x,y)$ in raw pixel units so that decay parameters control peak sharpness in absolute pixels. The ground truth density map is
\begin{equation}
    Y(x,y) = \frac{1}{D(x,y)^{(a \cdot D(x,y) + b)} + c}
\end{equation}
where $a, b, c$ are decay hyperparameters. Following \citet{delplanque2023herdnet}, we set $a=0.02$, $b=0.65$, $c=1.0$. Animal centres take the value $1$; the response decays rapidly toward zero, and we zero out values below $0.01$ to suppress residual background activations.

\subsection{The Overhead Wildlife Locator (OWL) Framework}
\label{subsec:architecture}

OWL is built on a modular encoder--decoder design. Three variants probe the trade-off between computational efficiency, global-context modelling, and feature quality from large-scale self-supervised pretraining:
\begin{enumerate}
    \item \textbf{OWL-C (Convolutional):} a fully convolutional architecture for rapid inference.
    \item \textbf{OWL-T (Transformer-augmented):} OWL-C extended with Swin Transformer refinement blocks injected into the feature pyramid.
    \item \textbf{OWL-D (Foundation-pretrained):} replaces the CNN encoder with a frozen vision foundation model paired with a DPT-style decoder.
\end{enumerate}

\subsubsection{CNN Backbone and Feature Pyramid (OWL-C / OWL-T)}
OWL-C and OWL-T use Deep Layer Aggregation (DLA-34) \citep{yu2018deep} as the backbone. DLA-34's hierarchical tree aggregation fuses multi-scale feature maps at every encoder stage, preserving fine spatial detail effectively --- a useful property for small targets in high-resolution imagery. The backbone extracts a feature pyramid $F = \{f_0, f_1, \dots, f_5\}$ at strides $\{1, 2, 4, 8, 16, 32\}$.

\subsubsection{Swin Transformer Feature Enhancement (OWL-T)}
To improve separation of animals from complex textures (rocks, scrub), OWL-T injects a multi-scale Swin Transformer module \citep{liu2021swin} into the feature pyramid. For a feature map $f_i$:
\begin{equation}
    f'_i = \begin{cases}
        f_i + \text{SwinBlock}(f_i) & \text{for OWL-T} \\
        f_i & \text{for OWL-C}
    \end{cases}
\end{equation}

Each SwinBlock combines Window-based Multi-head Self-Attention (W-MSA) and Shifted Window Multi-head Self-Attention (SW-MSA) over non-overlapping $M \times M$ windows ($M = 8$ at early scales and $M = 4$ at deep scales, selected empirically to balance compute and receptive-field coverage). The shifted window operation enables cross-window information exchange.

The depth of the refinement is tailored per pyramid level. We use $[0, 1, 2, 2, 2, 3]$ blocks for scales $0$ to $5$, allocating more attention-based computation to the low-resolution, semantic-rich layers while keeping high-resolution early layers lightweight.

\subsubsection{DINOv3 Vision Foundation Backbone (OWL-D)}
\label{subsubsec:dinov3}

OWL-D replaces the DLA-34 backbone with a frozen DINOv3 \citep{simeoni2025dinov3} ViT-H+/16 encoder. DINOv3 is the latest member of the DINO family of self-supervised vision foundation models \citep{caron2021emerging, oquab2024dinov2}, pretrained on approximately $1.7$ billion curated web images and regularized with a Gram-matrix consistency objective to maintain dense feature quality at high resolutions. This variant tests whether large-scale self-supervised representations transfer to small-target aerial wildlife localization more effectively than an ImageNet-pretrained, in-domain-fine-tuned supervised encoder (the DLA-34 of OWL-C/OWL-T).

The encoder splits each input into non-overlapping $16 \times 16$ patches and processes them through $D$ Transformer blocks. We extract intermediate token representations $z_0, z_1, z_2, z_3$ from four evenly spaced encoder layers, at zero-indexed block positions $\ell_k = \lfloor (k+1) \cdot D / 4 \rfloor - 1$ for $k \in \{0, 1, 2, 3\}$, where $D$ is the encoder depth. Because patch tokenization requires input dimensions divisible by $16$, we apply reflective padding on the fly and crop the output back to $H/d \times W/d$ with $d = 2$.

To recover spatial multi-scale structure from the single-resolution ViT feature stack, we adopt a DPT-style \citep{ranftl2021vision, chen2022vitadapter} \emph{Reassemble} module. Each extracted layer $z_k$ is projected to a common decoder dimension $C_{\text{dec}} = 256$ via a $1{\times}1$ convolution, then resampled to a target stride by transposed convolution at strides 4 and 8, identity at stride 16, and a strided convolution at stride 32. The resulting pyramid $\{R_0, R_1, R_2, R_3\}$ is merged bottom-up through four \emph{FeatureFusion} blocks. Each block refines the previous fused tensor $F_{k+1}$ with a Pre-Activation Residual Convolutional Unit, adds the refined tensor to the current reassembled feature $R_k$, refines the sum with a second residual unit, upsamples by a factor of two, and projects through a $1{\times}1$ convolution:
\begin{equation}
    F_k = \mathrm{C}\bigl(\mathrm{U}_{2\times}\bigl(\mathcal{P}(R_k + \mathcal{P}(F_{k+1}))\bigr)\bigr),
\end{equation}
where $\mathcal{P}$ denotes a Pre-Activation Residual Convolutional Unit (two $3{\times}3$ convolutions with ReLU and batch normalization, wrapped by an identity skip), $\mathrm{U}_{2\times}$ is bilinear $2\times$ upsampling, and $\mathrm{C}$ is a $1{\times}1$ convolution. The deepest block ($k = 3$) has no skip and applies only the outer $\mathcal{P}$, $\mathrm{U}_{2\times}$, and $\mathrm{C}$. The fused output at stride $2$ feeds the same localization head as OWL-C and OWL-T (Section~\ref{subsubsec:decoder}).

The 840.6M-parameter encoder is held frozen throughout training, so only the Reassemble module, the FeatureFusion blocks, and the localization head ($14.8$M trainable parameters) receive gradients --- roughly $57\times$ fewer trainable parameters than the encoder.

\subsubsection{Decoder and Localization Head}
\label{subsubsec:decoder}
For OWL-C and OWL-T, the feature pyramid $F' = \{f'_0, \dots, f'_5\}$ is aggregated by the DLAUp decoder \citep{yu2018deep}. DLAUp repeatedly merges adjacent pyramid levels: each step upsamples a deep feature map and fuses it with the next-shallower one through a $3{\times}3$ convolution, accumulating semantic information from deep layers into shallow, high-resolution outputs. The decoder restores resolution to $H/d \times W/d$ with $d = 2$. For OWL-D, the DPT FeatureFusion pyramid of Section~\ref{subsubsec:dinov3} replaces DLAUp at the same output stride. In all three variants the prediction head is a $3{\times}3$ convolution, a $1{\times}1$ convolution, and a Sigmoid activation that constrains $\hat{Y}$ to $[0, 1]$.

\subsection{Loss Function}
\label{subsec:loss}

We train the OWL models with the penalty-reduced pixel-wise focal loss of CenterNet \citep{zhou2019objects}, which handles the extreme class imbalance between sparse animal centres and the vast background:

\begin{equation}
\resizebox{1.0\columnwidth}{!}{%
    $\displaystyle \mathcal{L}_{loc}(\widehat{Y}, Y) = -\sum_{i}\sum_{j}
    \begin{cases}
    (1 - \widehat{y}_{ij})^\alpha \log(\widehat{y}_{ij}) & \text{if } y_{ij} = 1 \\
    (1 - y_{ij})^\beta (\widehat{y}_{ij})^\alpha \log(1 - \widehat{y}_{ij}) & \text{otherwise}
    \end{cases}$%
}
\end{equation}

Here $Y, \widehat{Y}$ are the ground-truth and predicted localization grids and $y_{ij}, \widehat{y}_{ij}$ are their values at pixel $(i,j)$. The condition $y_{ij} = 1$ applies exclusively to the centre pixel of each annotated animal (where the FIDT map peaks at unity); all other pixels carry continuous values in $(0,1)$ or zero after thresholding, and $(1 - y_{ij})^\beta$ reduces the penalty for false positives near object centres. We set the focal hyperparameters $\alpha = 2$ and $\beta = 4$.

\subsection{Inference and Counting}
\label{subsec:inference}

To recover discrete animal locations from $\hat{Y}$, we apply a $3\times3$ max-pooling: a pixel $(x,y)$ is a detection if it equals the local maximum and exceeds an adaptive threshold $T_{adapt} = 0.2 \times \max(\hat{Y})$. Relative thresholding reduces sensitivity to illumination variation across survey flights. $T_{adapt}$ operates on the raw density map.

\section{Experimental Setup}
\label{sec:exp}

\subsection{Datasets}
We evaluate OWL on five publicly available aerial datasets covering both fixed-wing and UAV platforms and a wide spectrum of animal density --- 5.9 to 56.2 animals per image (Table~\ref{tab:datasets}). The collection spans open savannas (Delplanque/HerdNet, Eikelboom), managed grassland and paddock environments (Livestock/Han, Cattle/Shao, SheepCounter). We adhere to the official train/test splits of each dataset.

\begin{table*}[h]
\centering
\caption{Summary of the aerial datasets used in our benchmark. To facilitate fair comparison, we report statistics for the \textbf{original full-resolution images} before patching. The counts are presented as \textit{Number (Percentage of Dataset)} to illustrate the balance of the train/test split.}
\label{tab:datasets}
\resizebox{\textwidth}{!}{%
\begin{tabular}{l l l l c c c c c c}
\toprule
\multirow{2}{*}{\textbf{Dataset}} & \multirow{2}{*}{\textbf{Species}} & \multirow{2}{*}{\textbf{Habitat}} & \multirow{2}{*}{\textbf{Platform}} & \multirow{2}{*}{\textbf{Avg. Res. (px)}} & \multicolumn{2}{c}{\textbf{Train}} & \multicolumn{2}{c}{\textbf{Test}} & \multirow{2}{*}{\textbf{Density (obj/img)}} \\
\cmidrule(lr){6-7} \cmidrule(lr){8-9}
& & & & & \textbf{Imgs (\%)} & \textbf{Objs (\%)} & \textbf{Imgs (\%)} & \textbf{Objs (\%)} & \\
\midrule
\textbf{Delplanque (HerdNet)} & Multi-species (African) & Savanna & Fixed-wing & $5743 \times 4013$ & 928 (78.2\%) & 6,962 (75.2\%) & 258 (21.8\%) & 2,299 (24.8\%) & 7.8 \\
\textbf{Eikelboom} & Multi-species (African) & Savanna & Fixed-wing & $5080 \times 3383$ & 393 (77.8\%) & 3,003 (77.9\%) & 112 (22.2\%) & 850 (22.1\%) & 7.6 \\
\textbf{Livestock (Han)} & Livestock & Paddock & UAV & $3840 \times 2160$ & 82 (92.1\%) & 4,501 (90.0\%) & 7 (7.9\%) & 500 (10.0\%) & 56.2 \\
\textbf{Cattle (Shao)} & Cattle & Grassland & UAV & $4000 \times 3000$ & 260 (79.3\%) & 1,730 (89.1\%) & 68 (20.7\%) & 212 (10.9\%) & 5.9 \\
\textbf{SheepCounter} & Sheep & Paddock & UAV & $640 \times 640$ & 6,044 (97.2\%) & 192,954 (97.2\%) & 174 (2.8\%) & 5,623 (2.8\%) & 32.0 \\
\midrule
\textbf{Total} & - & - & - & - & \textbf{7,707} & \textbf{209,150} & \textbf{619} & \textbf{9,484} & \textbf{-} \\
\bottomrule
\end{tabular}%
}
\end{table*}

\subsection{Data Preprocessing}
\label{subsec:preproc}
Original images often exceed 17 MP for the fixed-wing datasets, so we use a patch-based training strategy. We extract $512 \times 512$ patches with a 160-pixel sliding-window overlap to avoid object truncation at patch boundaries. The \textit{SheepCounter} dataset, whose images are natively $640 \times 640$, is treated specially: a single $512 \times 512$ crop per image is extracted, retaining only annotations inside the crop.

To handle class imbalance we apply dataset-specific background sampling. For the smallest dataset (Han, 82 training images) we retain 100\% of background patches to preserve data diversity. For Cattle/Shao, Eikelboom, and Delplanque ($\geq 260$ training images each) we randomly sample 50\% of background patches. SheepCounter is a special case: every source image contains sheep clusters, so the dataset offers no genuinely empty scenes; we therefore evaluate counting performance on the full annotated set and acknowledge that the model has limited opportunity during training to learn to predict zero in truly empty regions. All patches containing at least one animal are retained in every case. The final benchmark training pool contains 18{,}091 patches with 183{,}600 point annotations. This annotation total is lower than the 209{,}150 full-image training objects of Table~\ref{tab:datasets} because patch extraction does not preserve every annotation: the dominant effect is the \textit{SheepCounter} single-$512{\times}512$-crop policy, which discards annotations falling outside the retained crop (\textit{SheepCounter} alone contributes ${\sim}92\%$ of all training objects), with a smaller contribution from background subsampling; patches bearing at least one animal are always kept. 

OWL is trained with VerticalFlip and HorizontalFlip (each $p=0.5$); RandomRotate90 ($p=0.5$) for arbitrary aerial orientations; RandomBrightnessContrast (limits $\pm 0.2$, $p=0.2$) and Blur (kernel up to $15$~px, $p=0.2$) for illumination and image-quality variation; and ImageNet-statistics normalization (mean $=[0.485, 0.456, 0.406]$, std $=[0.229, 0.224, 0.225]$). Patching and augmentation are applied only during training; at evaluation, full-resolution images are processed with the strategies of Section~\ref{subsec:slicing}.

\subsection{Evaluation Metrics}
\label{subsec:metrics}

We evaluate with counting and detection metrics computed at the full-resolution image level after reconstructing predictions from patches.

\textbf{Counting metrics.} We report Mean Absolute Error (MAE) and Root Mean Squared Error (RMSE):
\begin{equation}
    \text{MAE} = \frac{1}{N} \sum_{i=1}^{N} |C_i - \hat{C}_i|, \quad \text{RMSE} = \sqrt{\frac{1}{N} \sum_{i=1}^{N} (C_i - \hat{C}_i)^2}
\end{equation}
where $N$ is the number of test images and $C_i, \hat{C}_i$ are the ground-truth and predicted counts for image $i$.

\textbf{Detection metrics.} A predicted point $\hat{p}$ is a True Positive (TP) for a ground-truth point $p$ if $||\hat{p} - p||_2 \leq \tau$, with $\tau = 40$ pixels throughout. A sensitivity analysis over $\tau \in \{20, 40, 60\}$ confirms that model rankings are stable (\ref{app:tau}). To enforce one-to-one matching we apply greedy nearest-neighbour assignment: pairwise Euclidean distances are computed, sorted, and assigned iteratively. Unmatched predictions are False Positives (FP), unmatched ground-truth points are False Negatives (FN). We report Average Precision (AP), and the area under the Precision--Recall curve. For density-based models (OWL, HerdNet), the peak activation value at each detected local maximum --- bounded in $[0, 1]$ by the Sigmoid output --- serves as the detection score. We also report the total predicted count and its signed percentage error relative to ground truth.

\textbf{Confidence intervals.} We compute bootstrap 95\% confidence intervals (CIs) with $B = 1{,}000$ image-level resamples. For counting metrics, per-image absolute and squared errors are recomputed on each resample. For detection metrics, per-image TP/FP/FN counts at every confidence threshold are summed over the resample, from which a Precision--Recall curve and its AP are derived. The 2.5th and 97.5th percentiles define the reported intervals. The counting threshold $t^*$ (Section~\ref{subsec:implementation}) is fixed at its test-set optimum across all bootstrap samples. Where within-OWL AP differences are small and the corresponding intervals overlap --- for example OWL-D vs.\ OWL-C on \textit{Cattle (Shao)} ($0.990$ vs.\ $0.988$) and OWL-T vs.\ OWL-C on \textit{SheepCounter} ($0.978$ vs.\ $0.977$) --- we treat the comparison as a statistical tie. We retain AP, which is threshold-free, as the primary ranking metric.

\subsection{Slicing Strategies}
\label{subsec:slicing}
Most test images greatly exceed the $512 \times 512$ training patch size, making direct inference on full-resolution images infeasible. We use tailored large-image inference per architecture. For the YOLOv11 family we apply Slicing Aided Hyper Inference (SAHI) \citep{akyon2022slicing}. The official POLO repository lacks native large-image support, so we implemented a custom sliding-window pipeline that runs inference on overlapping crops and merges outputs through a distance-based point non-maximum suppression. For OWL family and HerdNet, we adopt the continuous stitching pipeline of \citet{delplanque2023herdnet}: a sliding window with 160-pixel overlap averages density maps across overlapping regions, normalizing by the visit count to ensure smooth transitions across patch boundaries before local-maxima detection.

\subsection{Implementation Details}
\label{subsec:implementation}
All models are implemented in PyTorch and trained on a single NVIDIA A100-SXM4 (40~GB) GPU. Training uses the $512 \times 512$ patches of Section~\ref{subsec:preproc} for 30 epochs. No per-dataset hyperparameter tuning was performed.

HerdNet \citep{delplanque2023herdnet} is included with its official pre-trained weights from the original authors (trained on the Delplanque split) and is therefore evaluated only on the Delplanque test set, scored for localization through the same pipeline as the other models. We treat it as a published reference baseline rather than re-implementing it across the benchmark; the rationale for this choice is given in Section~\ref{subsec:specialization}.

For the fully supervised baselines we train YOLOv11n and YOLOv11l from the Ultralytics framework, initialized with COCO pre-trained weights (\texttt{yolo11n.pt}, \texttt{yolo11l.pt}). Training uses SGD with momentum $0.937$ and weight decay $5\times 10^{-4}$; the learning rate decays linearly from $0.01$ to $10^{-4}$ across the 30 epochs. Augmentation includes HSV jitter, random horizontal flip ($p=0.5$), and random translation and scale; mosaic augmentation is enabled throughout training except for the final 10 epochs.

For POLO \citep{may2024polo} we initialize with YOLOv8n (\texttt{yolov8n.pt}), SGD with momentum $0.937$ and weight decay $5\times 10^{-4}$, learning rate decaying linearly from $0.01$ to $10^{-4}$. Geometric hyperparameters follow the published POLO defaults except for the base pseudo-box radius (set globally to $24$ pixels, the average target size across all training patches), the Distance-over-Radius (\texttt{dor}) threshold ($0.5$), and the localization-loss multiplier (\texttt{loc}, set to $5$ to prioritize centre regression). We fix this radius globally rather than per dataset by design: no other method in the benchmark receives per-dataset parameter tuning under our matched, no-tuning protocol, so a per-dataset POLO radius would advantage POLO relative to the uniformly-treated baselines. 

OWL-C, OWL-T, and OWL-D are trained with the Adam optimizer, initial learning rate $1\times 10^{-4}$, weight decay $5\times 10^{-4}$, with the FIDT supervision and focal loss of Sections~\ref{subsec:fidt}~and~\ref{subsec:loss}.

\textbf{Backbone initialization.} The OWL-C/OWL-T DLA-34 backbone is ImageNet-pretrained and fine-tuned end-to-end; the OWL-D DINOv3 ViT-H+/16 encoder is self-supervised on LVD-1.7B and held frozen (Section~\ref{subsubsec:dinov3}); YOLOv11n/l are COCO-initialized and POLO is YOLOv8n-initialized. The benchmark therefore matches the \emph{training budget} across methods (30 epochs, no per-dataset tuning) rather than the pretraining corpus: ``fairness'' here means matched compute and epoch budget, not matched pretraining.

\textbf{Counting threshold selection.} For density-based models (OWL, HerdNet) the predicted count is obtained by thresholding the output map at a scalar confidence level $t \in [0,1]$; for detection-based models (YOLO, POLO) $t$ is the detection-score cutoff. For each model and dataset independently we select the threshold $t^{*}$ that minimizes MAE on the test images. Reported MAE and RMSE therefore represent the best achievable values at the optimal threshold. This post-hoc selection introduces an optimistic bias; because the threshold is selected independently per model the bias applies uniformly across methods, so relative comparisons remain valid. AP is unaffected, as it integrates over all thresholds.

\section{Results}
\label{sec:results}

\subsection{Quantitative Comparison}
\label{subsec:quantitative}

Table~\ref{tab:results} summarizes performance of the three OWL variants against POLO, HerdNet, and the YOLOv11 family across all five datasets. OWL-D records the highest AP on four of five conditions --- \textit{Delplanque (HerdNet)} (0.934), \textit{Eikelboom} (0.854), \textit{Livestock (Han)} (0.973), \textit{Cattle (Shao)} (0.990) --- while OWL-T leads on the extreme-density \textit{SheepCounter} UAV set (0.978), where OWL-D drops to 0.834. Across the three OWL variants, OWL records the lowest RMSE on four of five conditions: OWL-C on \textit{Delplanque} (2.93), OWL-D on \textit{Eikelboom} (2.32) and \textit{Livestock (Han)} (3.18), and OWL-T on \textit{SheepCounter} (2.08). On \textit{Delplanque}, OWL-C and OWL-T already exceed HerdNet (0.840) at AP 0.901 and 0.900; OWL-D adds a further 3.3 points to reach 0.934 (Section~\ref{subsec:specialization}). OWL-C retains the lowest RMSE on this dataset (2.93 vs.~OWL-D 3.44).

POLO's structural limitations are most visible on the free-ranging wildlife datasets. On \textit{Eikelboom}, POLO records the lowest AP of all models (0.591 (0.536, 0.651)), lagging 26 points behind OWL-D and 3 points behind YOLOv11n (the POLO--vs--YOLO and POLO--vs--OWL comparisons are partially confounded by backbone choice; see Section~\ref{subsec:density-vs-polo}). POLO is competitive only on \textit{SheepCounter}, where uniform backgrounds and consistent animal scale match its fixed-radius assumption.

OWL-D dominates on \textit{Eikelboom} across all three metrics --- AP 0.854 (0.806, 0.898), MAE 1.52 (1.19, 1.86), RMSE 2.32 (1.86, 2.78). The AP gain over YOLOv11l is large enough that their bootstrap intervals do not overlap; the gap over OWL-T is directionally consistent on all three metrics but the CIs partially overlap. It is the only condition in which one variant leads detection accuracy and both counting metrics at once.

On \textit{SheepCounter}, OWL-T and OWL-C retain the top AP (0.978 and 0.977) and the lowest counting error (OWL-T MAE 0.77, RMSE 2.08), while OWL-D drops to AP 0.834 (0.810, 0.856) with MAE 4.63 and RMSE 8.28 --- a 12.9\% population undercount examined in Section~\ref{subsec:sheep_tradeoff}. The bootstrap intervals for OWL-D and OWL-T do not overlap. The fully supervised YOLOv11l (AP 0.921) incurs higher counting error than OWL-C/T (MAE 2.17, RMSE 3.91), consistent with non-maximum suppression discarding valid detections in densely clustered scenes.

The two UAV datasets expose contrasting failure modes. On \textit{Livestock (Han)}, YOLOv11l incurs the highest counting error of all models (MAE 24.29, 18\% population undercount), while OWL-D records the lowest MAE 2.43 (1.00, 3.86) and RMSE 3.18 (1.65, 4.39) --- a roughly threefold reduction in RMSE relative to the next-best variant OWL-T (RMSE 9.75) and to YOLOv11n (RMSE 11.76). The small test set ($n = 7$) inflates the bootstrap intervals; the caveat is noted in Section~\ref{subsec:limitations}. On \textit{Cattle (Shao)}, OWL-D achieves the highest AP (0.990) with OWL-C close behind (0.988); counting performance converges across methods, with YOLOv11l recording the lowest MAE (0.16) and RMSE (0.40), though bootstrap intervals overlap substantially for all models. Scaling YOLOv11 from n to l on \textit{Eikelboom} raises AP from 0.622 to 0.736 but both are overtaken by OWL-D at 0.854 (a 12-point margin over YOLOv11l), and the same ranking holds for RMSE.

\begin{table*}[h]
\centering
\caption{Detection (AP) and counting (MAE, RMSE) performance at $\tau = 40$ pixels. All metrics include bootstrap 95\% confidence intervals ($B = 1{,}000$ image-level resamples) shown in {\scriptsize parentheses}. \textbf{GT} denotes the Ground Truth total count. The \textbf{Pred.} column shows the total predicted count (percentage error relative to GT). Best values per metric are highlighted in \textbf{bold}. $^\dagger$Pre-trained weights evaluated with our pipeline.}
\label{tab:results}
\resizebox{\textwidth}{!}{%
\begin{tabular}{l l c c c c}
\toprule
\textbf{Dataset (GT)} & \textbf{Model} & \textbf{AP} ($\uparrow$) & \textbf{MAE} ($\downarrow$) & \textbf{RMSE} ($\downarrow$) & \textbf{Pred. (Err \%)} \\
\midrule
\multirow{7}{*}{\shortstack[l]{\textbf{Delplanque (HerdNet)} \\ (GT: 2,299)}}
& HerdNet$^\dagger$ \citep{delplanque2023herdnet} & 0.840 {\scriptsize(0.800, 0.872)} & 1.66 {\scriptsize(1.36, 2.00)} & 3.05 {\scriptsize(2.42, 3.65)} & 2,198 (-4.4\%) \\
& POLO & 0.727 {\scriptsize(0.662, 0.788)} & 3.11 {\scriptsize(2.55, 3.73)} & 5.82 {\scriptsize(4.51, 7.10)} & 2,105 (-8.4\%) \\
& YOLOv11n & 0.782 {\scriptsize(0.741, 0.815)} & 2.51 {\scriptsize(2.03, 2.97)} & 4.70 {\scriptsize(3.68, 5.60)} & 2,091 (-9.0\%) \\
& YOLOv11l & 0.858 {\scriptsize(0.827, 0.885)} & 1.84 {\scriptsize(1.48, 2.26)} & 3.77 {\scriptsize(2.84, 4.69)} & 2,140 (-6.9\%) \\
& OWL-C (Ours) & 0.901 {\scriptsize(0.870, 0.926)} & 1.31 {\scriptsize(1.00, 1.67)} & \textbf{2.93} {\scriptsize(2.08, 3.74)} & 2,234 (-2.8\%) \\
& OWL-T (Ours) & 0.900 {\scriptsize(0.874, 0.922)} & 1.33 {\scriptsize(1.05, 1.68)} & 2.94 {\scriptsize(1.97, 3.86)} & \textbf{2,284 (-0.7\%)} \\
& OWL-D (Ours) & \textbf{0.934} {\scriptsize(0.902, 0.960)} & \textbf{0.96} {\scriptsize(0.62, 1.39)} & 3.44 {\scriptsize(1.33, 5.32)} & 2,332 (+1.4\%) \\
\midrule
\multirow{6}{*}{\shortstack[l]{\textbf{Eikelboom} \\ (GT: 850)}}
& POLO & 0.591 {\scriptsize(0.536, 0.651)} & 3.02 {\scriptsize(2.48, 3.58)} & 4.22 {\scriptsize(3.40, 4.98)} & 710 (-16.5\%) \\
& YOLOv11n & 0.622 {\scriptsize(0.565, 0.683)} & 2.30 {\scriptsize(1.88, 2.77)} & 3.45 {\scriptsize(2.67, 4.18)} & 804 (-5.4\%) \\
& YOLOv11l & 0.736 {\scriptsize(0.687, 0.782)} & 2.08 {\scriptsize(1.67, 2.53)} & 3.14 {\scriptsize(2.44, 3.83)} & 819 (-3.6\%) \\
& OWL-C (Ours) & 0.737 {\scriptsize(0.657, 0.808)} & 2.15 {\scriptsize(1.73, 2.61)} & 3.27 {\scriptsize(2.56, 3.96)} & \textbf{853 (+0.4\%)} \\
& OWL-T (Ours) & 0.755 {\scriptsize(0.680, 0.826)} & 2.01 {\scriptsize(1.64, 2.42)} & 2.95 {\scriptsize(2.37, 3.53)} & 795 (-6.5\%) \\
& OWL-D (Ours) & \textbf{0.854} {\scriptsize(0.806, 0.898)} & \textbf{1.52} {\scriptsize(1.19, 1.86)} & \textbf{2.32} {\scriptsize(1.86, 2.78)} & 834 (-1.9\%) \\
\midrule
\multirow{6}{*}{\shortstack[l]{\textbf{Livestock (Han)} \\ (GT: 500)}}
& POLO & 0.750 {\scriptsize(0.709, 0.794)} & 11.14 {\scriptsize(7.00, 15.57)} & 12.60 {\scriptsize(8.48, 16.37)} & \textbf{496 (-0.8\%)} \\
& YOLOv11n & 0.810 {\scriptsize(0.775, 0.853)} & 9.14 {\scriptsize(4.00, 14.86)} & 11.76 {\scriptsize(6.03, 16.57)} & 462 (-7.6\%) \\
& YOLOv11l & 0.689 {\scriptsize(0.574, 0.833)} & 24.29 {\scriptsize(12.14, 35.57)} & 28.73 {\scriptsize(18.43, 36.08)} & 410 (-18.0\%) \\
& OWL-C (Ours) & 0.868 {\scriptsize(0.848, 0.888)} & 10.43 {\scriptsize(4.43, 16.43)} & 12.97 {\scriptsize(6.73, 17.57)} & 515 (+3.0\%) \\
& OWL-T (Ours) & 0.818 {\scriptsize(0.779, 0.856)} & 8.14 {\scriptsize(4.14, 12.00)} & 9.75 {\scriptsize(5.88, 12.69)} & 481 (-3.8\%) \\
& OWL-D (Ours) & \textbf{0.973} {\scriptsize(0.956, 0.991)} & \textbf{2.43} {\scriptsize(1.00, 3.86)} & \textbf{3.18} {\scriptsize(1.65, 4.39)} & 507 (+1.4\%) \\
\midrule
\multirow{6}{*}{\shortstack[l]{\textbf{Cattle (Shao)} \\ (GT: 212)}}
& POLO & 0.870 {\scriptsize(0.809, 0.923)} & 0.75 {\scriptsize(0.54, 0.99)} & 1.17 {\scriptsize(0.91, 1.44)} & 191 (-9.9\%) \\
& YOLOv11n & 0.797 {\scriptsize(0.701, 0.898)} & 0.22 {\scriptsize(0.12, 0.32)} & 0.50 {\scriptsize(0.36, 0.63)} & \textbf{213 (+0.5\%)} \\
& YOLOv11l & 0.953 {\scriptsize(0.914, 0.986)} & \textbf{0.16} {\scriptsize(0.07, 0.25)} & \textbf{0.40} {\scriptsize(0.27, 0.50)} & 211 (-0.5\%) \\
& OWL-C (Ours) & 0.988 {\scriptsize(0.977, 0.996)} & 0.28 {\scriptsize(0.15, 0.43)} & 0.68 {\scriptsize(0.38, 0.99)} & 203 (-4.2\%) \\
& OWL-T (Ours) & 0.973 {\scriptsize(0.953, 0.991)} & 0.31 {\scriptsize(0.18, 0.46)} & 0.70 {\scriptsize(0.45, 0.92)} & 217 (+2.4\%) \\
& OWL-D (Ours) & \textbf{0.990} {\scriptsize(0.983, 0.997)} & 0.24 {\scriptsize(0.13, 0.35)} & 0.51 {\scriptsize(0.36, 0.65)} & 206 (-2.8\%) \\
\midrule
\multirow{6}{*}{\shortstack[l]{\textbf{SheepCounter} \\ (GT: 5,623)}}
& POLO & 0.953 {\scriptsize(0.944, 0.961)} & 1.72 {\scriptsize(1.40, 2.07)} & 2.91 {\scriptsize(2.33, 3.48)} & \textbf{5,567 (-1.0\%)} \\
& YOLOv11n & 0.919 {\scriptsize(0.909, 0.930)} & 1.92 {\scriptsize(1.57, 2.30)} & 3.20 {\scriptsize(2.49, 3.90)} & 5,563 (-1.1\%) \\
& YOLOv11l & 0.921 {\scriptsize(0.912, 0.930)} & 2.17 {\scriptsize(1.72, 2.64)} & 3.91 {\scriptsize(2.91, 5.03)} & 5,417 (-3.7\%) \\
& OWL-C (Ours) & 0.977 {\scriptsize(0.967, 0.986)} & 0.81 {\scriptsize(0.56, 1.11)} & 2.12 {\scriptsize(1.21, 2.99)} & 5,560 (-1.1\%) \\
& OWL-T (Ours) & \textbf{0.978} {\scriptsize(0.967, 0.987)} & \textbf{0.77} {\scriptsize(0.53, 1.10)} & \textbf{2.08} {\scriptsize(1.09, 3.13)} & 5,559 (-1.1\%) \\
& OWL-D (Ours) & 0.834 {\scriptsize(0.810, 0.856)} & 4.63 {\scriptsize(3.61, 5.66)} & 8.28 {\scriptsize(6.73, 9.91)} & 4,898 (-12.9\%) \\
\bottomrule
\end{tabular}%
}
\end{table*}

\subsection{Precision-Recall Analysis}
\begin{figure*}[h]
    \centering
    \includegraphics[width=0.65\textwidth]{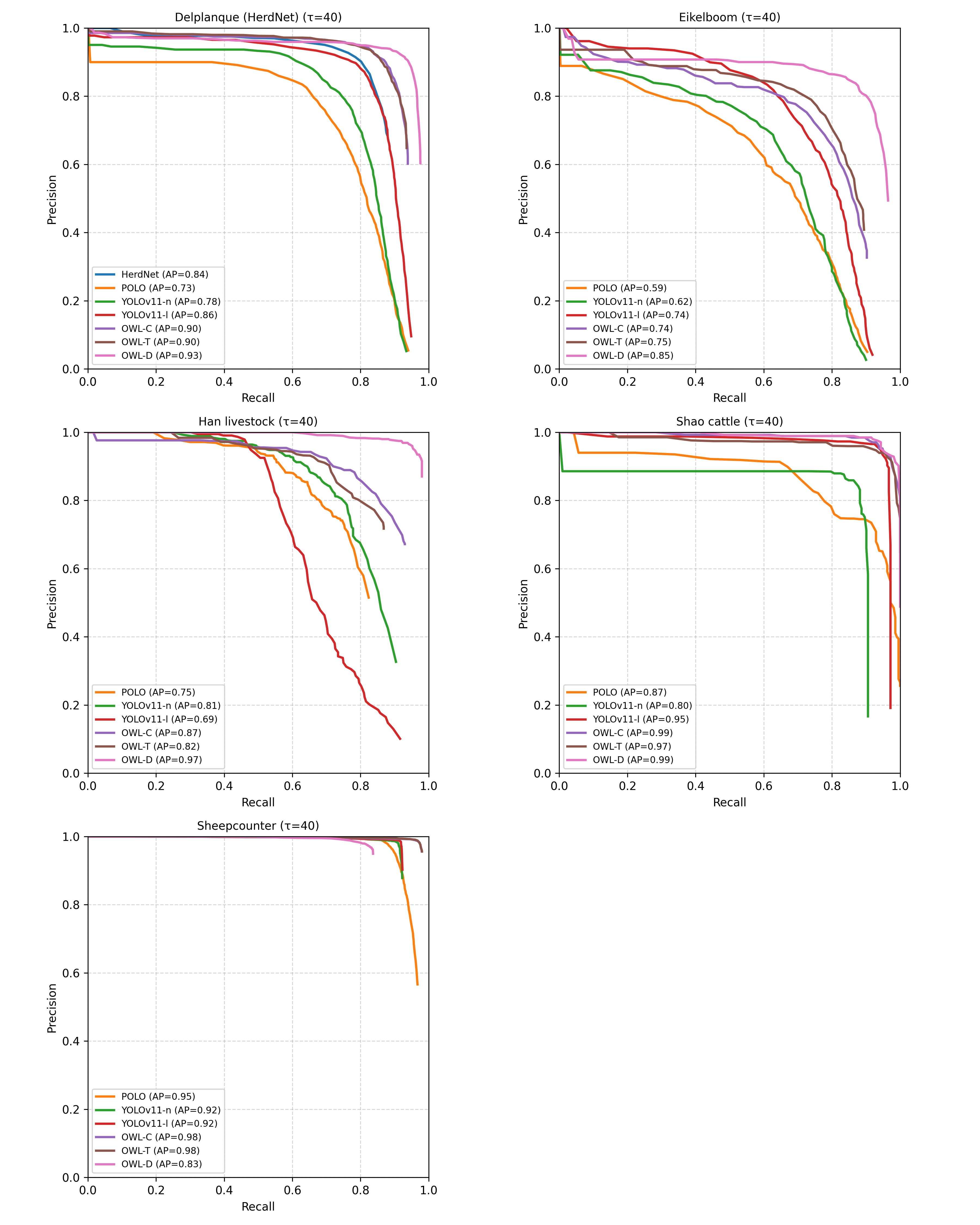}
     \caption{Precision--Recall curves at $\tau = 40$ pixels across the five benchmark datasets. OWL-D (pink) attains the highest AP on four of the five datasets (\textit{Delplanque}, \textit{Eikelboom}, \textit{Livestock (Han)}, \textit{Cattle (Shao)}), sustaining near-flat precision out to recall close to $1.0$. OWL-C (purple) and OWL-T (brown) are the next strongest, with similar high-recall behaviour. POLO (orange), YOLOv11n (green), and YOLOv11l (red) exhibit steeper precision drop-offs, most visibly on \textit{Eikelboom} and \textit{Livestock (Han)}. The ranking reverses on \textit{SheepCounter}: OWL-D collapses to $0.83$, while OWL-C, OWL-T, and the baselines preserve precision across the recall range. HerdNet (blue) is evaluated only on \textit{Delplanque}.}
    \label{fig:pr_curves}
\end{figure*}

To characterize model behaviour across the full precision-recall trade-off, we visualize the Precision-Recall (PR) curves in Figure~\ref{fig:pr_curves}.

Across all five datasets the three OWL variants share an extended high-precision plateau, retaining precision near the ceiling over a wide recall range and deferring their decline to higher recall than the baselines.

The separation is clearest on the free-ranging wildlife sets. On \textit{Delplanque} the OWL curves dominate HerdNet and the YOLOv11 family over essentially the entire range, with OWL-D extending furthest along the recall axis before its terminal drop; on the harder \textit{Eikelboom} set --- where no model sustains a flat plateau --- OWL-D is visibly detached from the field, mirroring its non-overlapping bootstrap interval (Table~\ref{tab:results}). The two UAV livestock sets contrast sharply: on \textit{Livestock (Han)} OWL-D traces a near-rectangular curve while YOLOv11l's precision collapses earliest and steepest (dropping below $0.5$ well before the others reach comparable recall, corroborating its MAE $24.29$), whereas on \textit{Cattle (Shao)} the regime is saturated and OWL-C, OWL-D, OWL-T, and YOLOv11l all hug the top-right corner with differences resolving only at the highest recall. On the extreme-density \textit{SheepCounter} set nearly all models stay near precision $1.0$ across most of the recall range; OWL-D is the lone exception, its curve falling away at a markedly lower recall --- the PR-domain signature of the $12.9\%$ undercount and reduced AP ($0.834$) examined in Section~\ref{subsec:sheep_tradeoff}, and the only condition in which OWL-D is recall-limited rather than recall-leading.

\subsection{Qualitative Comparison}

\begin{figure*}[h]
    \centering
    \includegraphics[width=\textwidth]{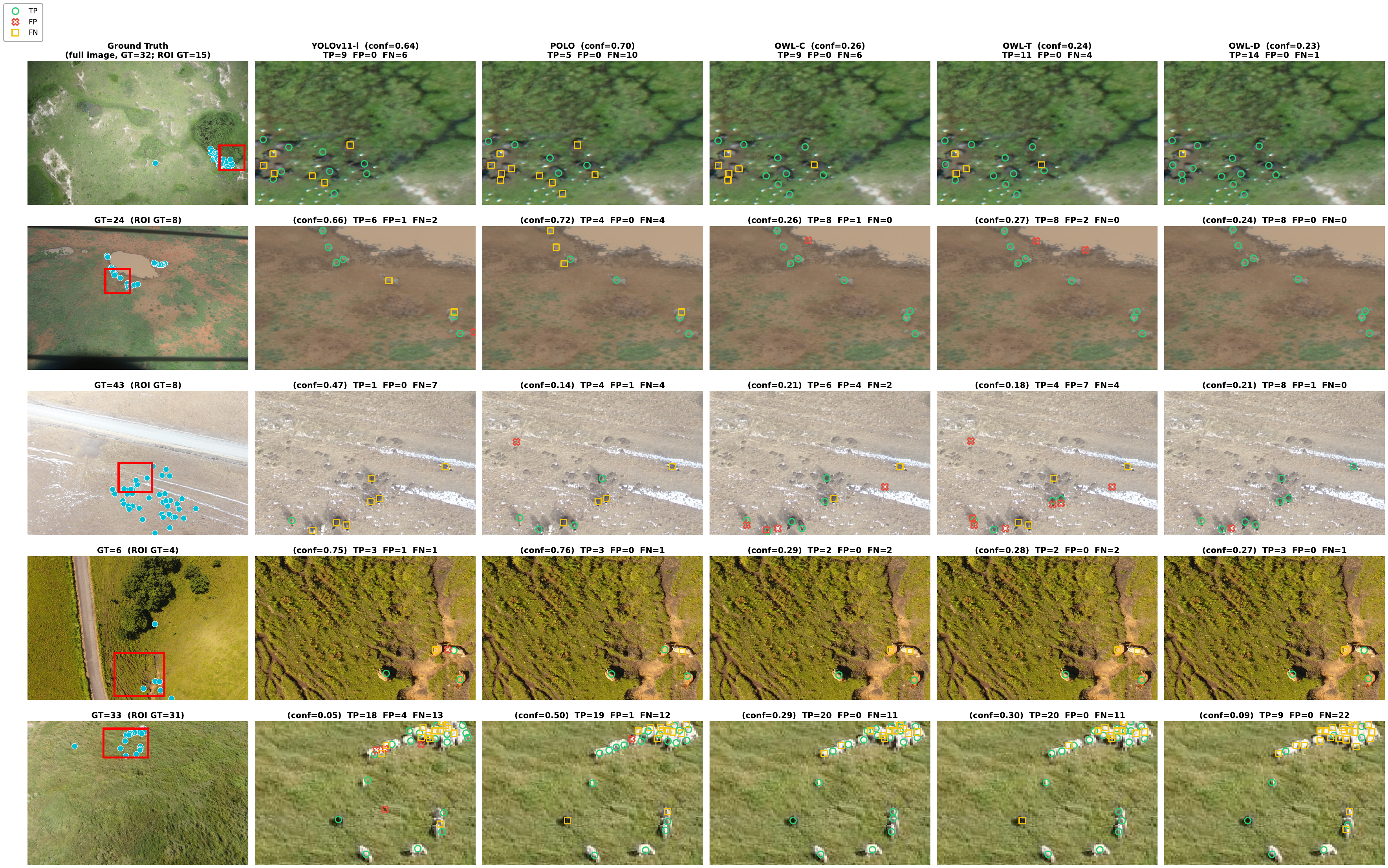}
    \caption{\textbf{Qualitative localization across density regimes.} Rows: fixed-wing savanna (\textit{Delplanque (HerdNet)}, Row 1; \textit{Eikelboom}, Row 2), UAV confined paddock (\textit{Livestock (Han)}, Row 3), sparse UAV cattle pasture (\textit{Cattle (Shao)}, Row 4), and dense UAV paddock (\textit{SheepCounter}, Row 5). Column~1: full-resolution image with point annotations (cyan dots) and the Region of Interest (ROI, red rectangle). Columns~2--6: predictions from YOLOv11l, POLO, OWL-C, OWL-T, and OWL-D restricted to the ROI at each model's per-dataset optimal confidence threshold (conf), matched greedily to GT at $\tau = 40$ px. Green circles: True Positives; red crosses: False Positives; yellow squares: False Negatives; per-cell titles report the (TP, FP, FN) triplet within the ROI.}
    \label{fig:qualitative}
\end{figure*}

Figure~\ref{fig:qualitative} compares model predictions across five density
 regimes. The Region of Interest (ROI) in each row is the red rectangle in
 Column 1. On the moderate fixed-wing savanna scenes (Rows 1--2), OWL-D
 delivers the cleanest detections: 14 of 15 animals recovered in the
 \textit{Delplanque} ROI (Row 1, 14 TP / 0 FP / 1 FN) and every annotation
 matched in the \textit{Eikelboom} ROI (Row 2, 8 TP / 0 FP / 0 FN). In Row 1,
 OWL-T is the closest baseline (11 TP / 4 FN), YOLOv11l and OWL-C each
 recover only 9 of 15, and POLO trails at 5 of 15. In Row 2, OWL-C and
 OWL-T also recover all 8 targets but at the cost of 1 and 2 false positives,
 respectively, while YOLOv11l recovers 6 of 8 (1 FP) and POLO 4 of 8 cleanly.
 On the UAV confined-paddock scene (\textit{Livestock (Han)}, Row 3, small
 densely clustered animals at high contrast), OWL-D maintains near-perfect
 recall (8 TP, 1 FP). OWL-T over-predicts in the same ROI (7 FP), while the
 detector baselines under-recover the herd (YOLOv11l: 7 FN). On the sparse
 UAV cattle pasture (Row 4), the detector baselines and OWL-D each recover
 3 of 4 animals with at most one false positive, while OWL-C and OWL-T slip
 to 2 of 4. The trend reverses on the dense UAV paddock scene
 (\textit{SheepCounter}, Row 5, 31 targets in the ROI): OWL-D recovers only
 9 of 31 (22 FN) despite operating at a reduced threshold (conf $=0.09$),
 while OWL-C and OWL-T preserve the most consistent localization, each
 recovering 20 of 31 cleanly (20 TP / 11 FN / 0 FP), with POLO close behind
 at 19 of 31 (1 FP) and YOLOv11l at 18 of 31 (4 FP). YOLOv11l keeps pace
 with POLO only by lowering its confidence threshold to $0.05$ --- compared
 with 0.64--0.66 on sparser datasets --- pointing to confidence-calibration
 drift for detection-based models under high density. These patterns confirm
 that OWL localizes more reliably than detector-based baselines across the
 full density spectrum, with the caveat that OWL-D's foundation-pretrained
 representation does not transfer to the extreme-density regime.

\subsection{Efficiency Analysis}
\label{subsec:efficiency}

To characterize the computational cost of each model, we benchmark inference efficiency on a common hardware platform. Table~\ref{tab:efficiency} compares the parameter count, theoretical complexity (GFLOPs), throughput, and observed latency on an NVIDIA A100 GPU. All measurements were conducted using a batch size of 1 on $512 \times 512$ inputs, averaging over 1,000 runs to ensure stability.

\begin{table*}[h]
\centering
\caption{Inference efficiency on NVIDIA A100. \textbf{Trainable} and \textbf{Total} report the parameter count seen during optimization and the full model size, respectively (these differ only for OWL-D, whose DINOv3 ViT-H+/16 encoder is frozen). \textbf{Latency} is mean inference time per patch; \textbf{P95} is the 95th-percentile latency; \textbf{Throughput} is patches per second.}
\label{tab:efficiency}
\begin{tabular}{l c c c c c c}
\toprule
\textbf{Model} & \textbf{Trainable} & \textbf{Total} & \textbf{GFLOPs} & \textbf{Latency} & \textbf{P95} & \textbf{Throughput} \\
& (M) & (M) & & (ms) & (ms) & (patches/s) \\
\midrule
YOLOv11n & \textbf{2.6} & \textbf{2.6} & \textbf{4.12} & 10.50 & 11.44 & 95.2 \\
YOLOv11l & 25.3 & 25.3 & 55.86 & 20.16 & 21.64 & 49.6 \\
POLO & 3.0 & 3.0 & 5.20 & \textbf{7.57} & \textbf{8.15} & \textbf{132.2} \\
OWL-C & 18.4 & 18.4 & 60.25 & 9.91 & 10.40 & 100.9 \\
OWL-T & 29.9 & 29.9 & 76.51 & 20.29 & 21.75 & 49.3 \\
OWL-D & 14.8 & 855.5 & 1681.01 & 201.56 & 202.57 & 5.0 \\
\bottomrule
\end{tabular}
\end{table*}

The results show a gap between theoretical complexity (GFLOPs) and real-world latency. OWL-C requires roughly $15\times$ more floating-point operations than YOLOv11n (60.25 vs 4.12 GFLOPs) but achieves a slightly lower mean latency (9.91 vs 10.50 ms). This is consistent with the known behaviour of depthwise separable convolutions on high-end hardware: such operations reduce parameter count but introduce memory-access overheads that limit parallelization, whereas the dense standard convolutions in OWL-C's DLA-34 backbone dispatch efficiently on the A100.

POLO is the fastest architecture (7.57 ms), benefiting from its YOLOv8n backbone, at the cost of the detection degradation in complex terrain noted in Section~\ref{subsec:quantitative}. YOLOv11l ($\approx 25.3$M parameters) exhibits an inference latency of 20.16 ms, almost identical to OWL-T (20.29 ms). The OWL-over-YOLO gain on most datasets is therefore not attributable to parameter capacity: at matched footprint and latency, the weakly supervised density-estimation framework still yields lower RMSE than fully supervised bounding-box detection. OWL-T's throughput ($\approx 49$ patches/s) is well within the operational requirements for offline survey analysis. OWL-D sits at the opposite end of the trade-off curve: although its trainable footprint ($14.8$M parameters) is actually smaller than every other OWL or YOLO variant we benchmark, the frozen $840.6$M DINOv3 ViT-H+/16 encoder must run on every forward pass, pushing mean latency to 201.56~ms (${\approx}10\times$ slower than OWL-T) and throughput to 5.0~patches/s. This precludes real-time use but remains tractable for batched offline processing of large-scale surveys: at this throughput, a 100{,}000-patch census --- equivalent to roughly $26$~gigapixels of imagery at $512 \times 512$ patch size --- completes in ${\approx}5.5$~hours on a single A100.

\section{Discussion}
\label{sec:discussion}

\subsection{The Economic Case: When Point Supervision is Sufficient}

The headline finding is that point-supervised density estimation can match or outperform fully supervised bounding-box detection regardless of detector parameter count. On four of five benchmark conditions OWL achieves lower RMSE than YOLOv11n; on the low-density \textit{Cattle (Shao)} set counting performance converges across methods within bootstrap intervals. This corroborates the emerging consensus that bounding boxes are not necessary for precision in aerial population surveys. On \textit{Delplanque}, for example, OWL-T reduces RMSE by over $37\%$ relative to YOLOv11n (2.94 vs 4.70) and keeps an edge against the higher-capacity YOLOv11l (2.94 vs 3.77), though bootstrap intervals partially overlap. Part of the gap reflects the inherent ambiguity of boxes in aerial imagery: clustered animals create box overlap that confuses the regression head, whereas points coupled with FIDT maps provide a sharp, unambiguous supervision signal. Given the reported $7\times$ \citep{ge2023point} and $3\times$ \citep{mullen2019comparing} annotation-cost advantages of points over boxes, OWL lets researchers annotate larger and more representative datasets for the same budget, yielding higher statistical power than smaller fully supervised datasets.

The scope of this claim must be qualified. Point supervision is sufficient for the \textit{counting and localization} objectives central to most population-level ecological surveys. Bounding boxes retain advantages for tasks that require the spatial extent of each individual --- morphometric analysis, body-condition scoring, and species classification from aerial imagery --- where the box encodes biologically meaningful geometric information that a point cannot. For surveys requiring such secondary analyses, a hybrid strategy --- points for large-scale screening and boxes for a targeted subset of ambiguous or scientifically critical cases --- may offer the most favourable trade-off.

Computational cost matters too. At matched footprint and latency (OWL-T vs.\ YOLOv11l; Section~\ref{subsec:efficiency}), the density-estimation framework still yields lower RMSE than fully supervised detection on most datasets --- with the \textit{Cattle (Shao)} exception and several differences within bootstrap intervals --- isolating the performance gap to the supervision paradigm rather than parameter capacity. In ecological monitoring, annotation time typically dominates compute time, so the labelling acceleration far outweighs the additional GPU inference cost; at $\approx 49$ patches/s OWL-T remains practical for large-scale survey processing. For practitioners, this validates OWL as a unified tool that maintains performance across the species, environments, and density regimes of our benchmark without architectural modification.

\subsection{Density Estimation vs.\ Pseudo-Box Detection}
\label{subsec:density-vs-polo}
The comparison with POLO highlights the advantage of pure density estimation over pseudo-box adaptations. POLO struggles on \textit{Eikelboom}, recording its lowest AP (0.591) across the benchmark. The likely cause is the rigidity of the fixed-radius assumption: a constant radius around each point works for uniform herds (e.g.\ sheep) but fails when animal scale varies or clutter is high. OWL's regression-based output creates a continuous probability landscape that is less sensitive to background clutter, suggesting that retrofitting a detector to accept points is less effective than an architecture intrinsically designed for density estimation. The comparison is partially confounded by backbone choice: POLO uses a YOLOv8n backbone (as released by its authors) while the fully supervised baselines use YOLOv11, so the POLO-vs-YOLO contrast cannot fully isolate the pseudo-box paradigm from generational backbone improvements. The same confound applies to OWL-vs-POLO, which differ in both supervision strategy and backbone (DLA-34 vs.\ YOLOv8n).

\subsection{Advancing the Baseline: Specialization and Encoder Choice}
\label{subsec:specialization}
OWL-D establishes a new state of the art on Delplanque (0.934 AP vs HerdNet's 0.840), despite HerdNet being trained and evaluated solely on this same dataset. Delplanque is among the most widely used aerial benchmarks for multi-species wildlife localization in African savanna environments, so this margin represents a meaningful advance on a reference condition for the field. The CNN-backbone OWL variants already clear this bar without foundation pretraining: OWL-C and OWL-T reach AP 0.901 and 0.900 respectively, exceeding HerdNet by 6.1 and 6.0 points, with OWL-D adding a further 3.3 points on top of OWL-C.

We attribute the improvement to two distinct factors. The first applies to all three OWL variants: OWL trains solely on the localization objective without a concurrent classification branch. HerdNet jointly optimizes density estimation and species classification, and these concurrent objectives may induce gradient conflicts that reduce localization precision. Decoupling them lets the encoder dedicate its full representational capacity to geometric fidelity. The $\sim$6-point gap between OWL-C/T and HerdNet at the same DLA-34 backbone supports this as the dominant factor. The second factor applies only to OWL-D: replacing the ImageNet-pretrained, in-domain-fine-tuned DLA-34 encoder with a frozen DINOv3 ViT-H+/16 introduces large-scale self-supervised foundation-model features that further improve dense prediction. Because OWL-D also replaces the decoder, we treat it as a distinct design point rather than an isolated encoder change (Section~\ref{subsec:design_points}).

\subsection{Three Architectural Design Points}
\label{subsec:design_points}
We compare three feature-extraction strategies under a shared FIDT supervision and LMDS inference recipe: a supervised CNN encoder (OWL-C), the same CNN encoder with Swin refinement (OWL-T), and a frozen DINOv3 ViT-H+/16 foundation encoder with a DPT-style decoder (OWL-D). OWL-C and OWL-T form a controlled pair --- identical ImageNet-pretrained DLA-34 backbone, DLAUp decoder, and training recipe, differing only in the injected Swin blocks --- so their contrast cleanly isolates that component. OWL-D, by contrast, changes both the encoder (frozen DINOv3 vs.\ DLA-34) and the decoder (DPT FeatureFusion vs.\ DLAUp), so its gains and its \textit{SheepCounter} collapse cannot be attributed to the encoder alone. Each strategy has a distinct empirical signature.

\textbf{OWL-C vs.\ OWL-T: peak sensitivity versus per-image variance.} CNNs achieve higher peak detection sensitivity while Transformer refinement produces more consistent per-image counts. The trade-off is clearest on \textit{Livestock (Han)}: OWL-C records a higher AP (0.868) than OWL-T (0.818) yet a substantially higher RMSE (12.97 vs 9.75). The AP/RMSE inversion is consistent with a model that is sensitive but not yet calibrated against high-confidence background activations; AP rewards any correctly matched detection regardless of per-image count variance, so OWL-C's high recall inflates its precision--recall area. Quantifying per-image false-positive (FP) counts at $\tau = 40$ shows that the difference is driven by per-image variance, not mean FP rate.\footnote{Per-image FP counts are computed at $\tau = 40$ pixels using the same greedy nearest-neighbour matching protocol as Table~\ref{tab:results}, with predictions thresholded at each model's $t^{*}$ (Section~\ref{subsec:implementation}).} On \textit{Livestock (Han)}, OWL-C and OWL-T produce nearly identical mean FP rates (13.29 vs 13.14 FP/image) but OWL-C exhibits substantially higher per-image variance ($\sigma = 7.7$ vs 5.5). On \textit{Eikelboom} OWL-C additionally averages more FPs per image (2.01 vs 1.49), consistent with OWL-T's lower RMSE (2.95 vs 3.27). The Swin blocks in OWL-T integrate scene-wide context to reduce per-image variance in complex terrain, yielding more consistent counts at the cost of a marginal AP reduction --- an ecologically meaningful trade-off where RMSE is the more relevant metric. On the low-density \textit{Cattle (Shao)} OWL-C leads OWL-T in both AP (0.988 vs 0.973) and MAE (0.28 vs 0.31), consistent with Transformer-based background suppression being most useful under density stress.

\textbf{OWL-D: foundation pretraining lifts AP on four of five datasets.} OWL-D records the highest point-estimate AP on four of five datasets --- its lead over the convolutional variants on \textit{Delplanque} and \textit{Cattle (Shao)} falls within bootstrap intervals (Section~\ref{subsec:metrics}) --- and the lowest counting error (MAE and RMSE) on \textit{Eikelboom} and \textit{Livestock (Han)} (Table~\ref{tab:results}). On \textit{Livestock (Han)} the absolute gain is largest: OWL-D reduces RMSE from OWL-T's 9.75 to 3.18, a ${\sim}3\times$ improvement that more than offsets the OWL-C/OWL-T variance trade-off above. On \textit{Eikelboom}, OWL-D is the only variant that simultaneously leads in mean AP, MAE, and RMSE; the AP advantage over YOLOv11l is supported by non-overlapping bootstrap intervals, while the advantage over OWL-T is directional and within bootstrap overlap. We interpret this as evidence that large-scale self-supervised pretraining --- with the Gram-regularized dense feature quality emphasized by DINOv3 \citep{simeoni2025dinov3} --- transfers effectively to sparse-to-moderate aerial wildlife scenes when paired with a DPT-style decoder. On the extreme-density \textit{SheepCounter}, OWL-D collapses (AP 0.834, MAE 4.63, RMSE 8.28), the only condition in which OWL-T retains the lead; we analyse this in Section~\ref{subsec:sheep_tradeoff}.

\subsection{Pretraining Distribution and the SheepCounter Trade-off}
\label{subsec:sheep_tradeoff}

The collapse of OWL-D on \textit{SheepCounter} --- AP $0.834$ and a $12.9\%$ population undercount, against AP $0.978$ and a $1.1\%$ undercount for OWL-T --- is the only condition in which the foundation-pretrained variant loses to the in-domain CNN variants. We report this gap as an open empirical observation and outline three non-exclusive hypotheses worth probing in future work.

First, DINOv3's pretraining corpus (LVD-1.7B, $\sim$1.7 billion curated web images) emphasizes dense feature quality at high resolutions but does not resemble nadir UAV paddock crops with $\sim$30--65 individuals per $512 \times 512$ image, asking a frozen representation to specialize to a regime substantially outside its training manifold. Second, ViT-H+/16 partitions each $512 \times 512$ input into $32 \times 32 = 1{,}024$ patch tokens of $16$ pixels each; on \textit{SheepCounter} a typical animal subtends only a handful of patches, so adjacent individuals share tokens and lose the within-patch texture the DPT decoder would need to resolve them --- a hard resolution floor that the stride-$2$ early layers of OWL-C/OWL-T do not impose. Third, the DINOv3 backbone is held frozen throughout training, so the encoder cannot adapt to the dense-counting regime; OWL-C and OWL-T train end-to-end and can specialize their representations to each dataset, while OWL-D's trainable components --- Reassemble, DPT fusion blocks, and localization head --- operate on top of a fixed feature distribution.

\subsection{Real-World Deployment: Alaskan Caribou Census}
\label{subsec:caribou_case_study}

To validate the operational viability of our framework, we deployed a fine-tuned OWL-C model in collaboration with the Alaska Department of Fish and Game to automate caribou counting from georeferenced aerial imagery. Historical survey data for this program consist entirely of legacy point-level annotations; bounding-box detectors are therefore not directly applicable without prohibitive re-annotation, illustrating the economic bottleneck of Section~\ref{sec:related}. We selected the lightweight OWL-C variant to prioritize high-throughput processing across vast terrains.

The model was fine-tuned on the publicly released 2017 Porcupine Caribou Herd (PCH) dataset, derived from 15 orthorectified GeoTIFF mosaics at GSD $\approx 3.6$~cm/pixel. The source mosaics were tiled into $512 \times 512$ patches with 160-pixel overlap, yielding 23{,}517 patches: 18{,}322 annotated patches with 273{,}268 point-level annotations and 5{,}195 background-only patches retained to improve specificity.

The deployed pipeline was applied to the full 2022 Central Arctic Herd (CAH) census --- a different herd captured five years later. The CAH 2022 survey comprises 44 georeferenced mosaics (24 with caribou annotations and 20 zero-GT mosaics retained as specificity controls), spanning 17.39 gigapixels (48.6~GB on disk) of contiguous Arctic terrain at GSD 2.2--4.0~cm/pixel with frame resolutions of up to $25{,}000 \times 25{,}000$ pixels. The same $512 \times 512$ / 160-pixel overlap tiling pipeline is applied at inference; OWL-C processes the resulting patches end-to-end on a single NVIDIA A100 GPU at the throughput reported in Table~\ref{tab:efficiency}, demonstrating practical feasibility for full-scale governmental surveys.

For reproducible evaluation, we report patch-level metrics on the publicly released CAH 2022 test subset --- 2{,}607 patches (1{,}852 annotated patches with 12{,}456 point-level annotations and 755 background-only patches) drawn from 19 source mosaics (of the 24 annotated CAH 2022 mosaics). At the matching threshold $\tau = 40$ image pixels used throughout the main benchmark (Table~\ref{tab:results}, \ref{app:tau}) and at the counting threshold $c^{*} = 0.20$, OWL-C attains $F_1 = 0.965$, recall $= 0.980$, precision $= 0.951$, and --- averaged over all 2{,}607 released test patches, including the 755 background-only patches --- MAE $= 0.27$ and RMSE $= 0.69$ caribou per patch, with a signed population-level error of $+3.1\%$ (12{,}840 predicted vs.~12{,}456 ground-truth annotations). We state plainly that $c^{*} = 0.20$ is selected, like every counting threshold in the main benchmark, to minimize error on the test patches themselves (Section~\ref{subsec:implementation}); it is therefore a \emph{test-optimal} operating point and carries the same bias we flag for the benchmark MAE/RMSE. We accordingly promote the threshold-robustness check from caveat to evidence: a confidence sweep shows $F_1$ varying by only $0.004$ across $c \in [0.18, 0.23]$, a flat plateau rather than a sharply tuned peak, so the reported $F_1$ is not an artifact of test-set selection and would survive a moderately mis-set threshold transferred from another survey. We further quantify empty-terrain specificity directly on the 755 background-only test patches ($0.198$ gigapixels of known-empty terrain), where every detection is by construction a false positive: at $c^{*} = 0.20$ OWL-C produces only 12 false detections --- $60.6$ per gigapixel --- with $98.7\%$ of background patches yielding no detection at all, and the few false positives confined to the $0.20$--$0.43$ confidence band (none high-confidence). This patch-level specificity is measured on background crops drawn from the same source mosaics as the annotated patches; the complementary deployment-scale test on the 20 zero-ground-truth whole-survey mosaics remains future work and would provide an even more stringent empty-terrain assessment (Section~\ref{subsec:limitations}). With that single scalar threshold the only quantity fit post-hoc, this remains a strict cross-herd and cross-temporal generalization test: the model was trained exclusively on PCH 2017 imagery and applied without any per-deployment retraining or weight/architecture tuning to a different herd (CAH), a different year (2022), and a different sensor configuration (PCH 50~mm lenses vs.~CAH mixed 50/90~mm camera array).

\begin{figure*}[h]
    \centering
    \includegraphics[width=0.6\textwidth]{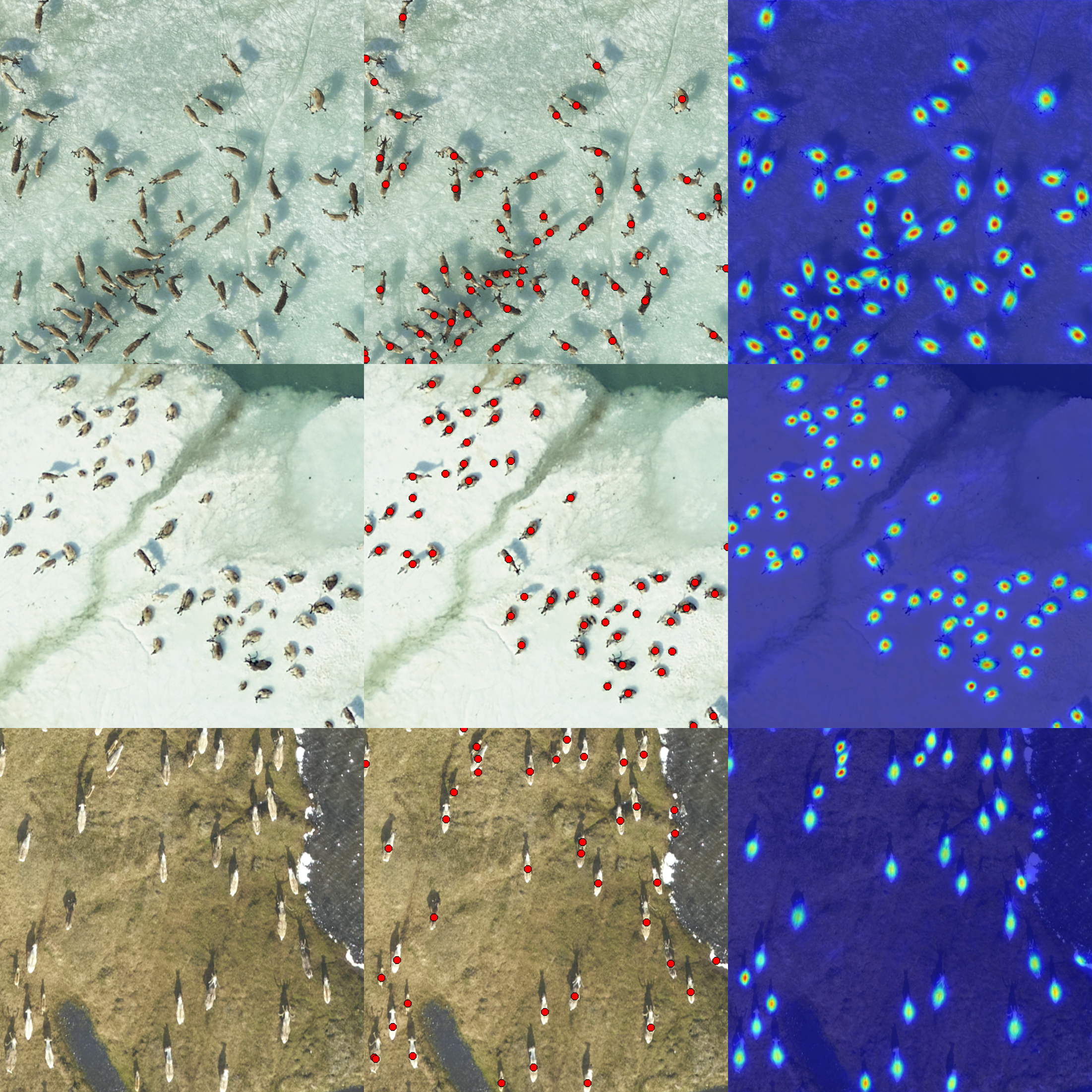}
    \caption{\textbf{Qualitative performance of the fine-tuned OWL-C on the Central Arctic Herd census.} Three representative patches from the deployment. Column~1: original aerial imagery. Column~2: model detections. Column~3: raw density heatmap overlaid on the source image. The model maintains distinct activation peaks even in tightly clustered formations (Rows~1 and~2); in highly heterogeneous terrain (Row~3) it occasionally produces false positives by confusing small isolated patches of snow near water and vegetation with caribou.}
    \label{fig:caribou_qualitative}
\end{figure*}

A qualitative assessment of the model's predictions on CAH is shown in Figure~\ref{fig:caribou_qualitative}. In the first two rows, OWL-C localizes the vast majority of the herd and preserves distinct density-map peaks even under tight clustering. The third row illustrates a representative failure mode in highly heterogeneous terrain: small isolated patches of melting snow along water and grass are occasionally misclassified as caribou, likely driven by their similar colour profiles, contrast, and apparent scale. Terrain-aware confidence calibration or post-hoc filtering of low-vegetation snow-water mixtures is a natural direction for further improvement.

This case study substantiates the core argument of the benchmark in practice: point-supervised density estimation is deployment-ready for governmental aerial wildlife monitoring, and the released patch-level evaluation reproduces this result on an independent herd and year.

\subsection{Limitations}
\label{subsec:limitations}

Five aspects of the evaluation merit consideration. First, the \textit{Livestock (Han)} subset contains only 7 original test images; results align with the broader benchmark trends but the wide bootstrap intervals limit confidence in this dataset --- a caveat that applies in particular to OWL-D's ${\sim}3\times$ MAE/RMSE improvement here. Second, several comparisons between OWL-T and YOLOv11l --- particularly on \textit{Delplanque} and \textit{Eikelboom} --- are directionally consistent but statistically inconclusive, as bootstrap intervals overlap. Third, the counting threshold $t^{*}$ is selected to minimize MAE on the test images rather than a held-out validation set, introducing an optimistic bias in the reported MAE and RMSE; per Section~\ref{subsec:implementation} this bias is applied uniformly across models and does not affect relative comparisons or AP. Fourth, OWL-D underperforms substantially on \textit{SheepCounter} (AP $0.834$, $12.9\%$ population undercount) despite leading the other four datasets. Fifth, empty-terrain specificity is quantified at the patch level --- $60.6$ false detections per gigapixel on the 755 background-only CAH test patches, with $98.7\%$ producing no detection (Section~\ref{subsec:caribou_case_study}) --- while the complementary deployment-scale test on the 20 zero-ground-truth whole-survey CAH mosaics remains future work and would provide the most stringent empty-terrain assessment.

\section{Conclusion}
\label{sec:conclusion}

This study addresses the scalability bottleneck in aerial wildlife monitoring: the cost of bounding-box annotation. We presented the Overhead Wildlife Locator (OWL), a weakly supervised framework that uses point-level labels and matches or exceeds fully supervised baselines (YOLO) and pseudo-box point-based methods (POLO).

OWL produces stronger and more consistent localization than pseudo-box methods. Even when compared against high-capacity bounding-box detectors (YOLOv11l) at virtually identical inference latency, the three OWL variants collectively record the lowest RMSE on four of five benchmark conditions (OWL-C on \textit{Delplanque}, OWL-D on \textit{Eikelboom} and \textit{Livestock (Han)}, OWL-T on \textit{SheepCounter}), with some pairwise differences between OWL-T and YOLOv11l within bootstrap intervals. OWL-D additionally posts the highest AP on four of five datasets --- decisively over the published HerdNet baseline on \textit{Delplanque} (bootstrap intervals disjoint) and over YOLOv11l on \textit{Eikelboom}, though its margin over the convolutional OWL variants on \textit{Delplanque} and \textit{Cattle (Shao)} is within bootstrap intervals --- with OWL-T remaining the strongest variant on the extreme-density UAV imagery where DINOv3 pretraining fails to generalize.

Operational readiness is further evidenced by the Alaskan Caribou Census deployment (Section~\ref{subsec:caribou_case_study}): on the publicly released CAH 2022 patch test set, OWL-C fine-tuned on the PCH 2017 training split achieves $F_1 = 0.965$ with a $+3.1\%$ population-level error under cross-herd and cross-temporal transfer.

The three architectural design points establish three complementary trade-offs across the OWL family: OWL-C offers the highest peak sensitivity and throughput ($\sim$100 patches/s) for rapid screening; OWL-T uses Swin refinement to reduce per-image FP variance in heterogeneous scenes and is the variant that best scales to extreme density; OWL-D delivers the strongest AP across four of five datasets and the lowest counting error on \textit{Eikelboom} and \textit{Livestock (Han)}, with an instructive failure on dense UAV paddock imagery.

These results support point supervision as a scalable alternative to bounding-box annotation for aerial population monitoring. The current OWL framework targets single-class localization and counting; extending it to multi-species classification --- via an auxiliary classification head trained alongside the FIDT-supervised localization branch, or via a two-stage pipeline that crops detected centres and routes them to a downstream classifier --- is a natural next step that would broaden the application range to multi-species census operations such as Delplanque-style savanna surveys. Other directions include diagnosing OWL-D's \textit{SheepCounter} failure through partial unfreezing and finer-patch ViT variants (e.g.\ ViT-S/14, ViT-B/14), optimizing OWL-C for real-time UAV edge inference where the OWL-D pipeline (5~patches/s at $\approx 1.7$~TFLOPs per patch) is impractical, extending OWL-T with temporal consistency for video-based tracking, and investigating cross-habitat transferability across diverse ecosystem types.

 \section*{Data Availability}
  We release two new annotated patch-level datasets from the Alaskan caribou aerial surveys:

  \begin{itemize}
      \item \textbf{PCH 2017 Training Set:} 23{,}517 image patches ($512 \times 512$ px) --- 18{,}322 annotated patches with 273{,}268 point-level annotations and 5{,}195 background-only patches --- from the 2017 Porcupine Caribou Herd survey (15 orthorectified GeoTIFF mosaics, GSD $\approx 3.6$~cm/px).
      \item \textbf{CAH 2022 Test Set:} 2{,}607 image patches ($512 \times 512$ px) --- 1{,}852 annotated patches with 12{,}456 point-level annotations and 755 background-only patches --- from the 2022 Central Arctic Herd survey (19 source mosaics, of the 24 annotated among 44 deployment mosaics, GSD 2.2--4.0~cm/px). The 20 zero-ground-truth mosaics used as specificity controls in the body of the paper (Section~\ref{subsec:caribou_case_study}) are not redistributed in patch form.
  \end{itemize}

  Both datasets, the OWL training and inference code, and the released model weights are available at \url{https://github.com/microsoft/MegaDetector-Overhead}. The five benchmark datasets used for model evaluation (Delplanque/HerdNet, Eikelboom, Livestock (Han), Cattle (Shao), SheepCounter) are publicly available from their respective original publications and are not redistributed here.


\section*{CRediT authorship contribution statement}
\textbf{Isai Daniel Chac\'on:} Conceptualization, Data curation, Formal analysis, Investigation, Methodology, Software, Validation, Visualization, Writing -- review \& editing.
\textbf{Zhongqi Miao:} Conceptualization, Data curation, Formal analysis, Investigation, Methodology, Software, Supervision, Validation, Writing -- review \& editing.
\textbf{Bruno Demuro:} Project administration, Writing -- review \& editing.
\textbf{Caleb Robinson:} Methodology, Writing -- review \& editing.
\textbf{Rahul Dodhia:} Resources, Supervision, Writing -- review \& editing.
\textbf{Lasha Otarashvili:} Conceptualization, Data curation, Methodology, Writing -- review \& editing.
\textbf{Jason Holmberg:} Conceptualization, Data curation, Methodology, Writing -- review \& editing.
\textbf{Kirk Larsen:} Conceptualization, Data curation, Methodology, Writing -- review \& editing.
\textbf{Howard Frederick:} Conceptualization, Data curation, Methodology, Writing -- review \& editing.
\textbf{Nathan J. Pamperin:} Conceptualization, Data curation, Methodology, Writing -- review \& editing.
\textbf{Pablo Arbel\'aez:} Conceptualization, Formal analysis, Investigation, Methodology, Supervision, Writing -- review \& editing.
\textbf{Juan M. Lavista Ferres:} Project administration, Resources, Supervision, Writing -- review \& editing.

\section*{Declaration of competing interest}
The authors declare that they have no known competing financial interests or personal relationships that could have appeared to influence the work reported in this paper.

\section*{Declaration of generative AI and AI-assisted technologies in the manuscript preparation process}
During the preparation of this work the authors used GitHub Copilot CLI in order to improve the language and readability of the manuscript and to assist with \LaTeX{} formatting and editing. After using this tool, the authors reviewed and edited the content as needed and take full responsibility for the content of the published article.

\section*{Acknowledgements}
We thank the Alaska Department of Fish and Game for providing the Porcupine Caribou Herd and Central Arctic Herd aerial survey imagery and field expertise, and Conservation X Labs, Kirk Larsen Consulting, and the Tanzania Wildlife Research Institute for their support throughout this work, including contributions to conceptualization, data curation, and methodology. We also thank the authors of the public benchmark datasets for making their data openly available.

\section*{Funding}
This research did not receive any specific grant from funding agencies in the public, commercial, or not-for-profit sectors.

\appendix
\renewcommand{\thesection}{Appendix~\Alph{section}}

\section{Sensitivity Analysis: Effect of the Matching Threshold $\tau$}
\label{app:tau}

Average Precision depends on the spatial distance threshold $\tau$ used to classify a predicted point as a True Positive. To validate the choice of $\tau = 40$ pixels used throughout the benchmark, we evaluate all models at $\tau \in \{20, 40, 60\}$ pixels. Counting metrics (MAE, RMSE) are excluded from this analysis, as they are determined by the confidence threshold rather than $\tau$.

\begin{figure*}[h]
    \centering
    \includegraphics[width=0.9\textwidth]{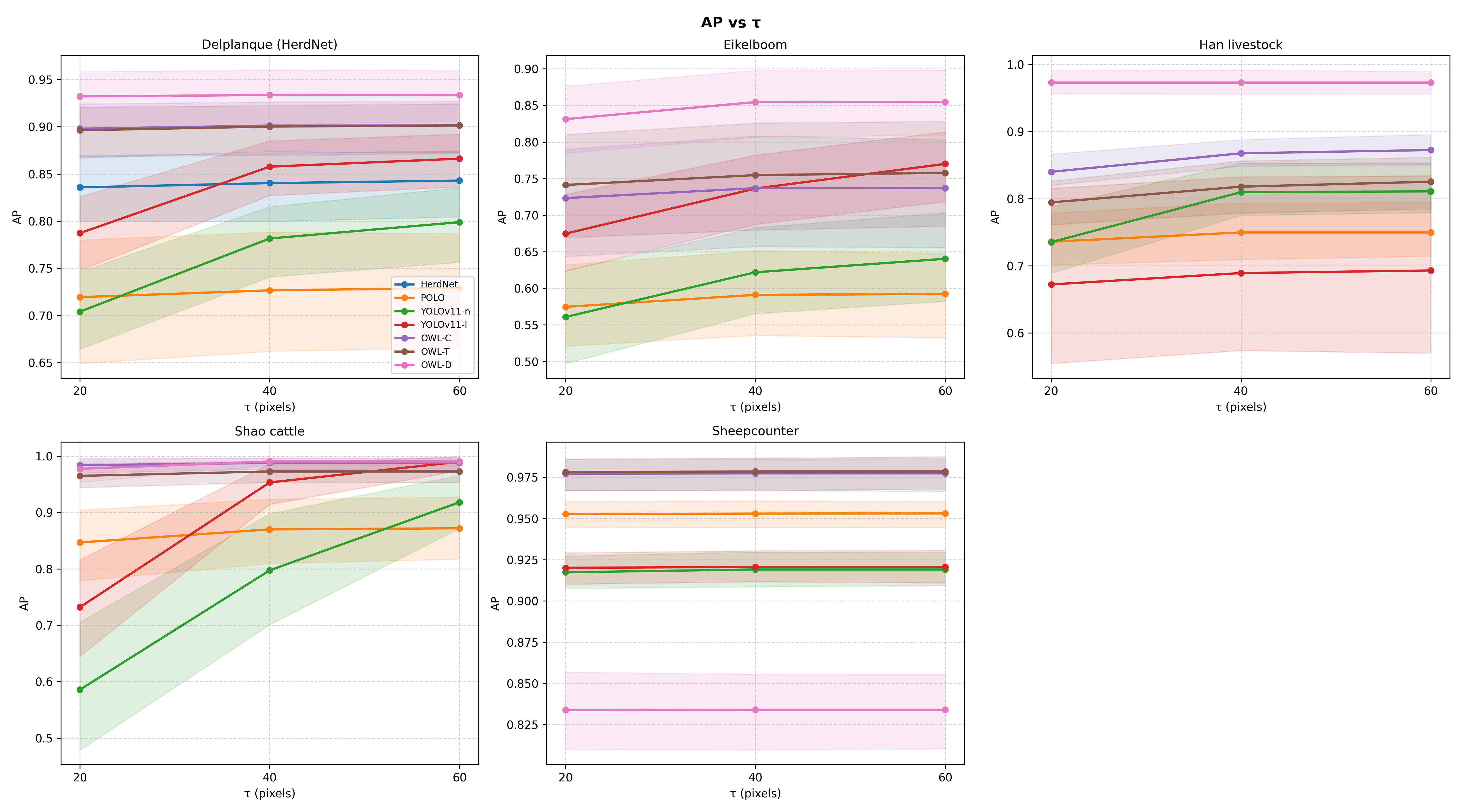}
    \caption{AP as a function of the matching threshold $\tau \in \{20, 40, 60\}$ pixels across the five benchmark datasets. OWL variants (purple/brown) exhibit minimal sensitivity, reflecting high localization precision. YOLOv11 variants (green/red) vary substantially on \textit{Cattle (Shao)} and, to a lesser extent, \textit{Eikelboom}, reflecting imprecise centre predictions under a strict matching criterion.}
    \label{fig:ap_tau}
\end{figure*}

Figure~\ref{fig:ap_tau} shows AP as a function of $\tau$ for all models across the five benchmark conditions, revealing two consistent patterns.

\textbf{OWL variants are insensitive to $\tau$.} Across all datasets, OWL-C and OWL-T AP changes by at most 0.003 from $\tau = 20$ to $\tau = 60$ (e.g.\ OWL-C on \textit{Delplanque}: 0.898, 0.901, 0.901). OWL-D is similarly stable on the four datasets where it leads --- AP varies by at most ${\sim}0.02$ across $\tau \in \{20, 40, 60\}$ (e.g.\ \textit{Eikelboom}: $\approx 0.831 \to 0.854 \to 0.855$; \textit{Cattle (Shao)}: $0.977 \to 0.990 \to 0.990$) --- and the OWL-D vs.\ OWL-T inversion on \textit{SheepCounter} is preserved across the full range. This reflects the high spatial precision of FIDT-supervised local maxima: predicted detection centres stay close to ground-truth annotations regardless of the matching window. POLO and HerdNet exhibit similarly low sensitivity on most datasets.

\textbf{Bounding-box models are sensitive to $\tau$ on low-density datasets.} The most pronounced sensitivity occurs on \textit{Cattle (Shao)}: YOLOv11n AP rises from 0.585 at $\tau = 20$ to 0.918 at $\tau = 60$, and YOLOv11l from 0.732 to 0.989. OWL-C changes from 0.983 to 0.988 over the same range. This indicates that bounding-box centre predictions are less precisely localized on the sparse grassland scene: valid detections exist but their centres are offset from the annotated point. A lenient $\tau$ recovers them; a strict one does not. The pattern is consistent with localization noise introduced by the box-regression head when targets are small and isolated.

The only dataset where rankings shift with $\tau$ is \textit{Eikelboom}: OWL-T leads YOLOv11l at $\tau \leq 40$ (0.755 vs 0.736 at $\tau = 40$), but YOLOv11l overtakes at $\tau = 60$ (0.770 vs 0.758) for the same reason. On all remaining datasets (\textit{Delplanque}, \textit{Livestock (Han)}, \textit{SheepCounter}), rankings are stable across the entire range.

We select $\tau = 40$ as the primary evaluation threshold: it avoids the over-strictness of $\tau = 20$, which can penalize valid detections marginally displaced by patch stitching, while remaining strict enough to prevent matching across adjacent animals in high-density scenes. At this threshold model rankings are consistent across all five datasets.

\bibliographystyle{cas-model2-names}

\bibliography{cas-refs}


\end{document}